Original Paper

# Algorithmic Reflectivity in Large Language Models: Methodological Study of Linguistic Structure in Mentalization

Stefano Epifani[1,2], MSc; Giuliano Castigliego[1,3], MSc; Giuliano Razzicchia[4], MSc; Elisabeth Seiwald-Sonderegger[5], MSc; Laura Kecskeméti[5], MSc

[1]Reseach Foundation, Digital Transformation Institute, Rome, Italy
[2]Department of Communication and Social Research (CoRiS), Sapienza Università di Roma, Rome, Italy
[3]Psychoanalytic Academy of Italian-Speaking Switzerland, Chur, Switzerland
[4]Digital Transformation Institute, Rome, Italy
[5]Psychiatric Services of the Canton of Grisons, Chur, Switzerland

**Corresponding Author:**

Stefano Epifani, MSc
Reseach Foundation, Digital Transformation Institute
Via Ottaviano, 42
Rome 00192
Italy
Phone: 39 3388810098
Email: stefano.epifani@sostenibilitadigitale.it

## Abstract

**Background:** Mentalization is a core human capacity involving interpretation of one's own and others' behavior in terms of underlying mental states. Within the mentalization-based treatment (MBT) framework, it is described along multiple dimensions, integrating cognitive, affective, relational, and regulatory processes. Large language models (LLMs) can generate linguistically reflective discourse, raising the question of whether the formal linguistic structure of mentalization can bbe reproduced without experiential and affective processes.

**Objective:** This study aimed to assess whether an LLM can generate mentalization profiles structurally coherent with MBT dimensions and recognized as formally mentalizing by expert clinicians. Specifically, it evaluated the structural and argumentative coherence of model-generated profiles and the level of agreement among MBT-trained psychiatrists. It also examined the distinction between psychological mentalization and algorithmic reflectivity, conceptualized as a formally coherent but nonexperiential linguistic phenomenon.

**Methods:** A comparative, descriptive methodological design was adopted. A total of 50 human-model dialogues were generated under standardized conditions. At the end of each interaction, the model produced a narrative mentalization profile structured along MBT dimensions. Five MBT-trained psychiatrists independently evaluated each profile using anonymized materials. Ratings covered 4 canonical MBT axes and 2 additional higher-order dimensions—regulatory functions and synthetic profile. For each dimension, they assigned separate Likert-scale ratings for evaluative and argumentative coherence. Descriptive statistics were computed at dialogue level. Interrater reliability was assessed using intraclass correlation coefficients (ICCs), primarily ICC(3,k), with CIs. The association between evaluative and argumentative coherence was examined using Pearson correlation coefficients with 95% CIs, with Spearman rank correlation as a robustness check.

**Results:** Across dimensions, mean scores ranged from 3.63 (SD 0.63) to 3.98 (SD 0.31), indicating moderate-to-high structural coherence. Interrater reliability was moderate to good, with ICC(3,k) values ranging from 0.60 to 0.84 in the evaluative domain. The highest scores were observed for implicit-explicit and synthetic profile, whereas internal-external showed the lowest mean. Score distributions showed limited variability and concentration in the upper range of the scale. Evaluative and argumentative coherence were strongly associated across all MBT dimensions, with Pearson $r$ ranging from 0.826 to 0.967 (all $P<.001$) and Spearman $\varrho$ ranging from 0.866 to 0.991. Informal qualitative comments frequently described the outputs as linguistically organized and clinically interpretable, while also noting limited affective variability and contextual integration.

**Conclusions:** The findings suggest that LLMs can reproduce the formal linguistic structure of reflective discourse, generating outputs perceived by clinicians as coherent and clinically interpretable across multiple dimensions of mentalization. However, within the present evaluative framework, this performance appears more consistent with structural language regularities than with evidence of experiential, affective, or intentional processes. These results support using LLMs as research and training tools for studying the linguistic aspects of mentalization, while highlighting the continuing importance of human clinical judgment in contexts requiring emotional, relational, and contextual integration.

# Introduction

Mentalization represents one of the core competencies of the human mind, that is, the ability to interpret one's own and others' behavior in terms of underlying mental states—intentions, desires, emotions, and beliefs [1].

This function integrates cognitive, affective, and relational dimensions, enabling individuals to assign meaning to subjective experience and to regulate their internal states within interpersonal contexts [2,3].

Within the mentalization-based treatment (MBT) framework, mentalization is commonly described along the following 4 canonical axes: cognitive-affective, internal-external, self-other, and implicit-explicit. In this study, 2 additional higher-order dimensions were also evaluated separately—regulatory functions and synthetic profile. In recent years, the emergence of large language models (LLMs) has reopened the debate on the nature of understanding and reflectivity.

Trained on massive text corpora and capable of generating coherent language, these systems appear able to reproduce forms of reasoning and dialogue analogous to human discourse [4,5]. Recent research has suggested that, under certain conditions, LLMs may display performance compatible with theory-of-mind tasks, showing an apparent capacity to represent others' mental states [6,7].

However, subsequent studies have shown that such results may depend heavily on the linguistic sensitivity of the model rather than on genuine psychological understanding, as even minor alterations in task wording can substantially affect performance [8].

Beyond clinical MBT literature, the study also relates to computational approaches to text evaluation, especially work distinguishing formal linguistic coherence from grounded understanding and previous attempts to operationalize reflective functioning through textual analysis.

Recent studies have reported that LLMs can generate responses that appear compatible with mentalizing tasks, including the personalization of emotional and mental state descriptions according to specific personality structures or psychopathologies [9]. However, these findings leave open the question of whether such performances reflect psychological mentalization or the formal linguistic organization of reflective discourse.

Do LLMs therefore *understand*, or do they simply *predict the linguistic form* of understanding?

From a clinical perspective, this question acquires particular relevance.

If mentalization is a function expressed through language, and if language itself possesses a reflective structure, then it is legitimate to ask to what extent a statistical model devoid of conscious experience can simulate the linguistic form of mentalization [1].

In other words, the issue is not whether an LLM "thinks," but whether it produces texts that appear mentalizing to a human observer.

This distinction recalls the difference proposed by Frith and Frith [10] between cognitive representations of mind and the phenomenological experience of intentionality.

MBT literature emphasizes that mentalization manifests not in the logical correctness of discourse but in its capacity to tolerate uncertainty, integrate perspectives, and acknowledge the limits of one's knowledge [1]. In this perspective, linguistic reflectivity and experiential reflectivity constitute complementary yet distinct levels.

The first is formalizable—and thus potentially simulable—as a set of syntactic and semantic rules; the second involves affective, embodied, and relational processes that lie outside the scope of statistical language [11,12].

However, analogous to caregivers' mirroring responses that help infants navigate chaotic emotional worlds, LLM-based psychotherapy could potentially reduce anxiety by identifying patterns, recognizing regularities, and rendering complexity comprehensible. The term "transformer" in models such as ChatGPT (OpenAI; generative pre-trained transformer) evocatively aligns with Bion's concept of transformation, highlighting their capacity to generate contextually suitable narratives to facilitate emotional and social learning. These tools may foster reflective thinking and problem-solving skills in certain populations [13].

From a computational perspective, however, it is important to situate these developments within the broader field of natural language processing (NLP). While existing work in computational text analysis has made significant progress in modeling affective and semantic features—such as sentiment analysis and the integration of textual and metadata signals to improve interpretative accuracy [14,15]—these approaches typically operate at a surface or mid-level representation of meaning.

More recent developments have further extended these approaches by integrating contextual and metadata signals and by leveraging LLMs for increasingly nuanced semantic representations [16-18]. For example, recent work has shown that combining linguistic features with contextual information can improve the detection of emotional tone and user intent [15].

However, such models generally remain limited to the classification or prediction of observable linguistic patterns and do not explicitly address the representation of underlying mental states, intentions, and epistemic positioning.

In contrast, this study adopts a clinically grounded framework based on mentalization theory, aiming to evaluate texts not only in terms of coherence or sentiment, but in terms of their capacity to represent self-other differentiation, integrate affect and cognition, and tolerate ambiguity.

From this perspective, the proposed evaluation dimensions can be understood as a higher-order extension of computational text assessment, bridging psychological theory and NLP by targeting latent mental-state representations rather than surface linguistic features.

Studying the linguistic behavior of an LLM in mentalization tasks therefore allows the formal linguistic dimension of reflective processing to be isolated, making observable what is implicit in the human mind—the grammar of reflective thought.

This study is situated in this frontier between language and mind.

Its aim is to evaluate whether, and to what extent, an LLM can generate mentalization profiles that are clinically coherent and recognized as plausible by expert observers.

Rather than assessing the model's ability to learn or correct itself, the study seeks to measure the structural and argumentative coherence of automatically generated mentalization profiles, comparing them with independent evaluations by MBT-trained psychiatrists.

Through a comparative analysis of a set of human-model dialogues, the study—beyond examining the LLM's ability to simulate mentalization processes—also offers empirical evidence addressing a crucial theoretical question: does mentalization emerge at a psychological level, or already at a linguistic one?

The results presented here suggest that reflectivity can be at least partially reproduced as a linguistic regularity, opening new avenues for research on the computational representation of the mental and its clinical and epistemological implications.

# Methods

## Study Design

The study adopts a comparative-descriptive design aimed at exploring the capacity of an LLM to generate mentalization profiles consistent with the theoretical and clinical parameters of MBT [1,3]. To ensure methodological transparency and reproducibility, the complete prompt architecture and inference settings used during data generation are reported in Multimedia Appendix 1. Because different phases of the interaction (story generation, dialogue, and profile generation) required different levels of variability, phase-specific inference parameters were used rather than a single global configuration.

Each model turn was executed as a separate stateless API request, with the phase-specific decoding settings reapplied in the request payload at each turn. The study protocol did not specify a fixed seed.

The primary objective is to assess the structural coherence and perceived clinical plausibility of the profiles generated by the model, by comparing them with the independent evaluations of 5 psychiatrists (including 3 authors—GC, ESS, and LK) trained in MBT.

In contrast to previous studies examining the use of LLMs in psychological contexts [4,6], this research does not investigate the model's capability to solve cognitive tasks but its competence in reproducing the linguistic form of reflective discourse.

The study is therefore cross-sectional and exploratory, involves no experimental manipulations or learning interventions applied to the LLM, and seeks to measure the stability and convergence of clinical evaluations assigned to the generated profiles.

The unit of analysis is the complete dialogue between the model and the human participant.

The study protocol was designed to include between 8 and 12 question-answer exchanges. In the final dataset, dialogue length ranged from 6 to 12 exchanges, as 2 sessions ended prematurely before reaching the planned minimum while still providing sufficient material for profile generation. At the end of the interaction, the model generated a free-text mentalization profile structured around the mentalization axes and grounded in the participant's responses. The quality of the profile was subsequently evaluated along the 4 MBT axes—cognitive-affective, internal-external, self-other, and implicit-explicit—together with regulatory functions and synthetic profile as 2 additional rated dimensions [2].

## Materials and Dialogue Generation

The analyzed corpus consists of 50 human-model dialogues conducted using GPT-4.1 (OpenAI), accessed through the OpenAI API using the model identifier "gpt-4.1." The model was configured to assume the role of a therapist engaged in a mentalization-oriented interview.

The term "human-model dialogues" is used to refer to interactions between participants and the language model, while the term "simulated" applies only to the model-generated story prompts initiating each dialogue.

Each dialogue was based on one of 20 stories generated by the model and subsequently validated by psychiatrists trained in MBT.

Validation was based on expert clinical judgment. Psychiatrists assessed thematic coherence, narrative plausibility, and the balance between cognitive and affective components to ensure consistency with stimuli commonly used in mentalization-oriented clinical settings.

No formal scoring system or exclusion criteria were applied during validation, and no stories were revised or discarded. All 20 stories were retained in their original form.

Stories were randomly assigned across the 50 dialogues without predefined balancing constraints. As a result, individual stories may have been used multiple times across different participants. This approach reflects the exploratory nature of the study and was intended to avoid introducing selection bias in the generation of dialogic material, while ensuring a baseline level of semantic consistency across stimuli.

For each dialogue, the model was provided with one of the stories and instructed to formulate a sequence of mentalization-oriented questions designed to explore the core MBT dimensions [1]. The exact phrasing of the prompts and instructions used to guide the model is reported in Multimedia Appendix 1 to ensure transparency and replicability.

The answers to these questions were provided by 15 human participants recruited through nonclinical convenience sampling; each participant contributed on average to 3 dialogues. No demographic or psychometric data were collected, as the aim was not to evaluate participants' mentalization quality, but to assess the model's ability to generate coherent and clinically interpretable mentalization profiles from heterogeneous input.

At the end of each dialogue, the model produced a narrative mentalization profile synthesizing the emerging content and organizing it according to the 4 MBT axes, together with the regulatory and synthetic dimensions.

These profiles were then submitted for evaluation by an independent group of expert clinicians.

## Participants and Raters

### Overview

A total of 5 clinical psychiatrists, all MBT-certified and with more than 10 years of therapeutic experience, served as raters.

All materials provided for evaluation were fully anonymized and included only the complete text of the dialogues (questions and answers) and the corresponding narrative profiles generated by the model. Texts were presented in a uniform format and contained no references to participant identity or to the origin of the profiles.

Raters were informed that they would evaluate anonymized narrative profiles derived from simulated dialogues. They were not told whether the texts were human-written, AI-generated, or mixed in origin. However, no formal manipulation check was conducted; therefore, the possibility that stylistic cues influenced assumptions about text origin cannot be excluded.

A formal posttask manipulation check would have strengthened interpretability. Such data were not collected in this study; we therefore explicitly acknowledge this as a limitation and identify it as a priority for future research.

All interactions between human participants and ChatGPT-4.1, the resulting narrative profiles, and the clinical evaluations performed by the psychiatrists were conducted natively in Italian. The system prompt architecture was designed and calibrated in Italian (Multimedia Appendix 1), ensuring that the clinicians evaluated the original linguistic nuances without translation biases.

Each of the 50 dialogues was examined by all clinicians, with individually randomized presentation order to reduce potential sequence effects, familiarity phenomena, or learning processes.

Evaluations were conducted independently, asynchronously, and without time constraints.

### Evaluation Procedure

For each dialogue, clinicians assigned 2 distinct Likert scores (1-5) for each of the six rated dimensions: (1) cognitive-affective, (2) internal-external, (3) self-other, (4) implicit-explicit, (5) regulatory functions, and (6) synthetic profile.

Two distinct dimensions of judgment were assessed for each axis:

- Evaluative coherence: the degree to which the generated profile was consistent with the clinician's own interpretation of the dialogue (answering the question: "Is the profile consistent with your clinical evaluation of the dialogue?")
- Argumentative coherence: the internal logical consistency and grounding of the model's reasoning (answering the question: "Are the arguments provided by the model internally coherent and logically grounded?")

Both dimensions were rated separately on a 5-point Likert scale (1=very low coherence, 5=very high coherence).

In addition to these ratings, raters filled in a brief free-form descriptive comment aimed at qualifying the quality of mentalization expressed in the text.

## Data Analysis

Data analysis followed a descriptive and comparative approach, aimed at assessing the coherence and interclinician stability of the evaluations provided by the 5 psychiatrists. Given the exploratory nature of the study, analyses focused on measures of central tendency, dispersion, and concordance rather than on formal hypothesis testing.

In line with the study design, the individual dialogue was the unit of analysis. Scores were averaged across raters for each dialogue for axis-level descriptive statistics and correlation analyses (n=50), whereas Table 1 summarizes the distribution of individual rater scores. This approach isolated variability attributable to differences between profiles and reduced artificial inflation of precision arising from nonindependent observations.

**Table 1.** Distribution of individual rater scores.

| Score | Evaluative coherence | Argumentative coherence |
|---|---|---|
| 1 | 0.93% | 0.93% |
| 2 | 3.07% | 2.8% |
| 3 | 21.33% | 19.2% |
| 4 | 57.2% | 60.87% |
| 5 | 17.47% | 16.2% |

For each rated dimension (cognitive-affective, internal-external, self-other, implicit-explicit, regulatory functions, and synthetic profile), mean scores and SDs were calculated separately for evaluative coherence and argumentative coherence. These included mean, SD, SE ($SE=SD/\sqrt{N}$), median, and IQR, providing complementary indicators of central tendency and dispersion. Given the bounded and discrete nature of Likert-scale data, distributional properties were interpreted with particular attention to variability and concentration of scores.

The association between evaluative coherence and argumentative coherence was examined at the dialogue level using Pearson correlation coefficient ($r$), with 95% CIs. To account for the ordinal nature of Likert-scale ratings and potential deviations from normality, Spearman rank correlation coefficient ($\varrho$) was also computed as a nonparametric robustness check. In cases of null variance, correlation coefficients were not computed.

To assess potential effects of dialogue structure, the number of exchanges per dialogue was recorded and its association with coherence scores was examined using Spearman rank correlation coefficient.

In addition, a negative control condition was implemented to assess the specificity of clinician ratings with respect to dialogue-profile alignment. A subset of dialogue-profile pairs was deliberately mismatched by associating each dialogue with a profile generated from a different interaction. This manipulation allowed evaluation of whether coherence ratings were driven by generic linguistic properties of the profiles or by their correspondence with the originating dialogue.

Interrater agreement was estimated using the intraclass correlation coefficient (ICC), adopting multiple models to provide a comprehensive assessment of reliability. Specifically, ICC(3,k) was calculated to estimate the reliability of mean ratings across the 5 clinicians under a 2-way mixed-effects model with consistency definition, appropriate when raters are considered fixed [19]. ICC(3,1) was additionally computed to estimate single-rater reliability, while ICC(2,k), based on a 2-way random-effects model, was used to assess generalizability to a broader population of similarly trained raters.

CIs were calculated for all ICC estimates, enabling interpretation according to established guidelines [20].

The sample size of 50 dialogues was evaluated in relation to estimation precision. The reduction in SE with increasing sample size showed diminishing returns beyond approximately 40 observations, indicating stability of descriptive estimates. This pattern is consistent with methodological work on ICC precision, which suggests that moderate sample sizes (30-50 units) are generally sufficient to achieve stable reliability estimates [21]. Accordingly, the final sample was considered adequate for the exploratory and descriptive aims of the study.

The number of 5 independent raters was selected based on observed stability of ICC estimates across subsets of raters, with additional raters producing only marginal improvements. This choice reflects a balance between statistical robustness and feasibility.

Qualitative comments provided by clinicians were analyzed using a descriptive, nonsystematic approach aimed at identifying recurrent patterns in evaluative language. No formal qualitative framework was applied, as the primary objective of the study was quantitative assessment of coherence and interrater agreement.

Accordingly, qualitative observations reported in the Results and Discussion sections should be interpreted as informal clinical impressions rather than as findings derived from a formal qualitative analytic procedure.

## Ethical Considerations

This study did not involve clinical interventions, vulnerable populations, or the collection of identifiable personal or health-related data. Human participants were involved exclusively in providing written responses to structured dialogic prompts for research purposes, without the administration of psychological tests, diagnostic procedures, or therapeutic activities.

No demographic, clinical, or sensitive personal information was collected. All textual materials used for analysis were fully anonymized before evaluation. Participants were informed about the nonclinical and research-oriented nature of the study and provided voluntary consent to the use of their anonymized textual responses for research purposes.

The clinician raters evaluated anonymized textual outputs and were not informed about the origin of the profiles. However, no formal manipulation check was conducted to determine whether raters inferred authorship from stylistic cues.

The study does not fall within any of the categories for which the favorable opinion of an ethics committee is specifically required by current Italian and European legislation. In particular, it does not constitute a clinical trial of a medicinal product within the meaning of Regulation (EU) No 536/2014, Legislative Decree June 24, 2003, no. 211, and the Decree of the Ministry of Health of 30

November 2021; nor a clinical investigation on medical devices under Regulation (EU) 2017/745; nor an observational study on medicinal products under the Determinazione del Presidente AIFA n. 425/2024 of 8 August 2024 ("Linea Guida per la classificazione e conduzione degli studi osservazionali sui farmaci").

Participants were informed about the purpose and nature of the study before participation, and consent was obtained through online explicit request.

To ensure participant privacy, all text processed through the external commercial API was managed under strict data-handling agreements. Specifically, data privacy options were configured so that all submitted content was completely excluded from the provider's model training pipelines.

# Results

## General Description of the Evaluations

All 50 dialogues generated by the language model were independently evaluated by the 5 psychiatrists trained in MBT.

Mean scores assigned across the 6 rated dimensions ranged from 3.63 (SD 0.63) to 3.98 (SD 0.31), consistently within the mid-to-high range of the 1-5 scale. The highest values were observed in the implicit-explicit and synthetic dimensions (mean 3.98, SD 0.31), whereas the internal-external axis showed the lowest mean (mean 3.63, SD 0.63).

SDs ranged from 0.18 to 0.63, indicating limited dispersion across dialogues and suggesting a high degree of structural consistency in the generated profiles. Lower dispersion was observed in the implicit-explicit and cognitive-related dimensions, while comparatively higher variability emerged in the internal-external axis.

Distributional analyses confirmed a strong concentration of scores in the upper range of the scale. Median values were equal to 4 across all dimensions, while dialogue-level IQRs ranged from 0.0 to 0.6 and were 0.2 in most cases, indicating high convergence in central tendency. SEs ranged from 0.025 to 0.089, reflecting stable estimation precision.

Given the bounded and discrete nature of Likert-scale ratings, dispersion was interpreted primarily through SD and IQR. These indicators consistently showed limited variability across dimensions. Dialogue-level score distributions are illustrated in Figure 1, whereas Table 1 summarizes the distribution of individual rater scores.

Overall, approximately 76% (1884/2500) of ratings were ≥4, whereas approximately 4% (116/2500) were ≤2, confirming a strong concentration in the upper range and a high degree of consistency in clinician evaluations.

Interrater reliability was moderate to good across evaluative coherence axis, with ICC(3,k) values ranging from 0.60 to 0.84. CIs for ICC estimates are reported in Table 2.

**Figure 1.** Distribution of evaluative and argumentative coherence scores across mentalization-based treatment (MBT) dimensions. Boxplots show median, IQR, and distribution spread. Readers should note the limited dispersion and the clustering of scores in the upper range of the scale across dimensions, indicating high consistency in clinician evaluations and relatively stable perceived coherence of the generated profiles.

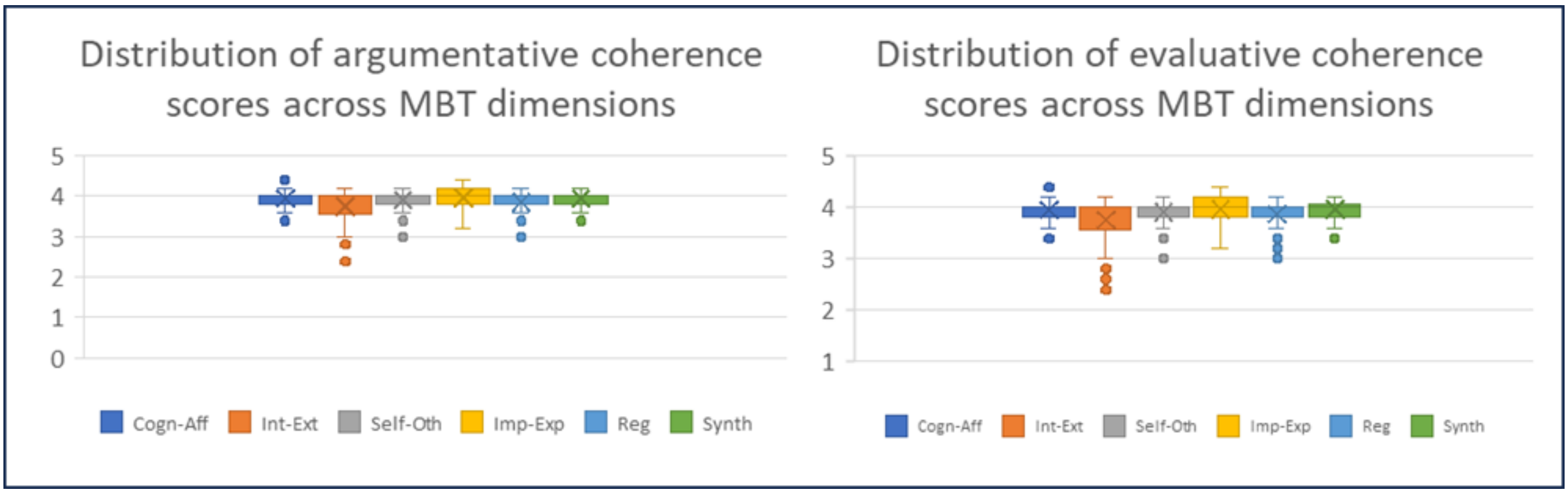

**Table 2.** Descriptive statistics and interrater reliability with 95% CIs.

| Factors and dimensions | Mean (SD) | SE | Median (IQR) | ICC(3,k)[a] (95% CI) |
|---|---|---|---|---|
| Cognitive - affective | | | | |
| Evaluative coherence | 3.83 (0.33) | 0.047 | 4 (3.6-4.0) | 0.65 (0.47 to 0.78) |
| Argumentative coherence | 3.88 (0.30) | 0.042 | 4 (3.8-4.0) | 0.58 (0.37 to 0.74) |
| Internal - external | | | | |
| Evaluative coherence | 3.63 (0.63) | 0.089 | 4 (3.4-4.0) | 0.84 (0.75 to 0.90) |
| Argumentative coherence | 3.71 (0.61) | 0.086 | 4 (3.6-4.0) | 0.83 (0.74 to 0.89) |

| Factors and dimensions | Mean (SD) | SE | Median (IQR) | ICC(3,k)[a] (95% CI) |
|---|---|---|---|---|
| Self - other | | | | |
| Evaluative coherence | 3.92 (0.29) | 0.041 | 4 (3.8-4.0) | 0.60 (0.39 to 0.75) |
| Argumentative coherence | 3.92 (0.27) | 0.038 | 4 (3.8-4.0) | 0.54 (0.30 to 0.72) |
| Implicit - explicit | | | | |
| Evaluative coherence | 3.98 (0.31) | 0.044 | 4 (3.8-4.0) | 0.69 (0.53 to 0.81) |
| Argumentative coherence | 3.98 (0.18) | 0.025 | 4 (3.8-4.0) | 0.24 (–0.15 to 0.53) |
| Regulatory functions | | | | |
| Evaluative coherence | 3.88 (0.30) | 0.042 | 4 (3.8-4.0) | 0.78 (0.67 to 0.86) |
| Argumentative coherence | 3.88 (0.31) | 0.043 | 4 (3.8-4.0) | 0.79 (0.67 to 0.87) |
| Synthetic profile | | | | |
| Evaluative coherence | 3.98 (0.26) | 0.037 | 4 (4.0-4.2) | 0.64 (0.46 to 0.78) |
| Argumentative coherence | 3.96 (0.22) | 0.031 | 4 (4.0-4.0) | 0.62 (0.43 to 0.77) |

[a]ICC: intraclass correlation coefficient.

Dialogue length ranged from 6 to 12 exchanges. Although the protocol targeted 8-12 exchanges, 2 dialogues ended prematurely after 6 exchanges while still yielding sufficient material for profile generation. The observed distribution was 6 exchanges (n=2, 4%), 8 exchanges (n=4, 8%), 9 exchanges (n=15, 30%), 10 exchanges (n=19, 38%), 11 exchanges (n=8, 16%), and 12 exchanges (n=2, 4%); no dialogues contained 7 exchanges. Correlation analyses showed no meaningful association between the number of exchanges and coherence scores (evaluative: $\varrho=-0.092$; argumentative: $\varrho=-0.058$), indicating that variation in dialogue length did not significantly influence clinician ratings.

In the negative control condition, where dialogues were paired with noncorresponding profiles, clinicians consistently assigned minimal coherence ratings (scores between 1 and 2 across dimensions). This pattern indicates a marked loss of perceived coherence when the alignment between dialogue content and profile interpretation is disrupted. These findings provide an empirical lower bound for coherence ratings and support the interpretation that clinician judgments are sensitive to dialogue-profile correspondence rather than to general linguistic fluency or structural regularity of the generated text.

## Convergence Among Raters

Convergence among the 5 psychiatrists was assessed using the ICC under different model specifications. Given that the study relies on the average ratings of 5 evaluators to determine clinical validity, the primary index reported is ICC(3,k), corresponding to a 2-way mixed-effects model with a consistency definition, appropriate for estimating the reliability of mean ratings.

Analyses were conducted separately for evaluative and argumentative coherence. Given the strong association observed between these dimensions (Spearman $\varrho$), they can be interpreted as capturing closely related aspects of the same underlying construct. However, interrater reliability was primarily examined in relation to evaluative coherence, as it more directly reflects clinical judgment.

Within the evaluative cluster, ICC(3,k) values ranged from 0.60 to 0.84, corresponding to moderate-to-good agreement according to commonly used ICC interpretation guidelines [20]. This pattern suggests a substantial level of convergence among clinicians in assessing the structural coherence of the generated profiles.

Argumentative coherence showed more variable ICCs across dimensions. In particular, the low ICC(3,k) for implicit-explicit argumentative coherence (0.24) likely reflects restricted between-dialogue variance (SD 0.18) and upper-range clustering, rather than greater interpretative variability among clinicians.

To assess the robustness of these findings, ICC(2,k) was also computed under a 2-way random-effects model with absolute agreement, enabling generalization to a broader population of similarly trained raters. As expected, ICC(2,k) values were lower, reflecting stricter agreement criteria. ICC(3,1) values were also computed to estimate single-rater reliability, confirming that agreement increases substantially when ratings are aggregated. The ICC(3,1) and ICC(2,k) estimates and their corresponding 95% CIs are reported in Multimedia Appendix 2.

Overall, these findings indicate that interrater agreement is stronger and more stable in the evaluative domain, supporting its role as the primary dimension for assessing the structural coherence of model-generated profiles.

## Axis-by-Axis Analysis

The axis-specific analysis is based on evaluative coherence scores, which more directly capture clinicians' judgments of profile consistency. Results are presented below by axis.

### Implicit-Explicit Axis

This dimension received the highest mean scores (mean 3.98, SD 0.31) among the MBT axes. Clinicians consistently noted the model's ability to make complex mental processes explicit, linking emotions, cognitions, and behaviors in a structured manner.

At the same time, several evaluations highlighted a tendency toward overstructured reasoning and reduced variability in narrative expression.

### Self-Other Axis

This dimension showed consistently high scores (mean 3.92, SD 0.29). Clinicians reported that the model effectively distinguished subjective perspectives from those of others, frequently using appropriate mental-state attribution forms (eg, "he may think that…," and "the other person might feel that…").

Evaluations highlighted the clarity and consistency with which perspectives were differentiated. At the same time, several clinicians noted that such differentiation remained primarily descriptive rather than dynamically elaborated.

### Cognitive-Affective Axis

Scores (mean 3.83, SD 0.33) indicated a general balance between cognitive and affective components, with a predominance of cognitive structuring. Clinicians reported that the model tends to rationalize emotional processes, representing them in structured and predictable forms.

Several evaluations noted that emotional content was consistently organized but showed limited variability in expression.

### Internal-External Axis

This axis showed the lowest mean scores (mean 3.63, SD 0.63) among the MBT dimensions. Generated profiles tended to describe internal states and external contexts in parallel, without integrating them into a dynamic or causal structure.

Clinicians noted that, although the discourse remained coherent, it was often decontextualized, with limited connection between subjective experience and ongoing events or relational dynamics.

### Regulatory Functions

Clinical evaluations indicated overall coherence in this dimension (mean 3.88, SD 0.3), but frequently highlighted superficiality in the representation of regulatory processes.

Regulatory passages were often described as prescriptive or overly linear, with limited variability in how emotional and behavioral dynamics were expressed.

### Synthetic Profile

This dimension showed among the highest mean scores (mean 3.98, SD 0.26). The model demonstrated a strong capacity to integrate heterogeneous information into coherent synthesis, producing stable and interpretable narrative structures.

Raters consistently noted the stability and clarity with which mental material was organized, although some evaluations indicated limited variability in how different elements were combined.

## *Coherence and Validity of Clinical Evaluations*

Clinicians frequently described the model-generated profiles as "coherent," "clinically plausible," or "logically structured."

However, recurring observations pointed to affective neutrality and limited emotional depth.

Although articulate and formally accurate, many profiles were informally described by clinicians as displaying an ordered form of reflectivity that appeared less affectively nuanced and experientially dynamic than typical human mentalization.

Evaluative coherence (the degree to which the model's profile matched the clinician's assessment) and argumentative coherence (the internal logic of the model's reasoning) showed a strong positive association, suggesting that clinicians integrate these dimensions during judgment. This relationship was quantitatively confirmed across all MBT dimensions using Pearson correlation coefficient calculated at the dialogue level (ie, based on mean scores aggregated across raters for each dialogue), with consistently high values ($r$=0.826-0.967, all $P$<.001) and narrow 95% confidence intervals, indicating a stable and robust association. Spearman's rank correlation coefficients showed a consistent pattern (ϱ=0.866-0.991), supporting the robustness of the observed relationship. Detailed correlation coefficients and CIs are reported in Table 3.

**Table 3.** Association between evaluative and argumentative coherence across mentalization-based treatment dimensions.

| Axis | Pearson correlation coefficient, *r* (95% CI) | Spearman rank correlation | *P* value |
|---|---|---|---|
| Cognitive - affective | 0.877 (0.792-0.929) | 0.866 | <.001 |
| Internal - external | 0.967 (0.942-0.981) | 0.986 | <.001 |
| Self - other | 0.918 (0.860-0.953) | 0.991 | <.001 |
| Implicit - explicit | 0.893 (0.818-0.938) | 0.986 | <.001 |
| Regulatory functions | 0.956 (0.924-0.975) | 0.953 | <.001 |
| Synthetic profile | 0.826 (0.711-0.898) | 0.924 | <.001 |

Texts perceived as logically coherent were also deemed clinically coherent, consistent with the link between syntactic order and perceived cognitive reflectivity described by Frith and Frith [10].

The very high associations observed between evaluative and argumentative coherence may reflect multiple factors. First, the 2 rating dimensions are conceptually related, as both involve judgments about the internal consistency and plausibility of the generated profiles. Second, because both ratings were assigned simultaneously by the same clinicians, shared evaluative tendencies may have contributed to the strength of the observed correlations. In addition, correlations were computed on dialogue-level aggregated scores, which may reduce variability across observations. Finally, the concentration of ratings in the upper range of the scale and the limited score dispersion across several dimensions may have contributed to the consistently high correlations observed across all dimensions. These findings therefore suggest that evaluative and argumentative coherence capture closely related aspects of perceived profile quality rather than fully independent constructs.

Quantitatively, the highest means in the implicit-explicit (3.98) and self-other (3.92) axes indicate that the model appears able to reproduce the grammar of intersubjective perspective-taking.

Conversely, lower scores on the internal-external axis (3.63) confirm the model's difficulties in linking subjective experience with environmental context, reflecting a tendency toward parallel rather than integrative descriptions [8].

### Summary of Findings

The integrated analysis of quantitative and qualitative data shows that the mentalization profiles generated by the model display high structural coherence, significant interpretative stability, and good clinical recognizability.

Mean scores between 3.63 and 3.98, moderate SDs, and evaluative ICC values between 0.60 and 0.84 suggest that the generated texts meet the formal criteria of reflective discourse, appearing readable, orderly, and internally consistent. In this sense, the model demonstrates the ability to organize language according to syntactic and semantic regularities that clinicians recognize as reflective. Profiles were coherent not only at the linguistic level, but also in distinguishing perspectives, articulating logical connections, and integrating narrative elements into structured interpretations. However, results suggest that observed coherence should not be interpreted as evidence of affective or intentional processes.

Although the model represents mental states formally, clinicians frequently perceived the generated profiles as less affectively dynamic and experientially nuanced than typical human mentalization.

These qualitative impressions should be interpreted cautiously, as they were derived from informal clinician comments rather than from a formal qualitative coding procedure.

Affective and contextual components—especially in the internal-external and partly in the cognitive-affective axes—remain predominantly descriptive rather than experientially grounded.

These findings support the interpretation of the model's output as a form of linguistic, rather than experiential, reflectivity, consistent with the notion of algorithmic reflectivity. In this perspective, reflectivity appears grammatical—coherent, repeatable, structurally organized—but lacking psychological depth.

## Discussion

### Principal Findings

The observed patterns indicate that, within the present experimental design, model-generated profiles are consistently perceived by clinicians as structurally coherent and clinically interpretable. This convergence suggests that generated outputs were perceived by clinicians as aligning with the formal criteria of narrative mentalization, particularly in terms of internal organization, logical structure, and interpretability. Rather than being interpreted as evidence of genuine mentalizing processes, these results point to the model's ability to reproduce the linguistic form of reflective discourse. In this sense, the observed coherence appears to derive from the organization of syntactic and semantic patterns that clinicians recognize as indicative of reflective thinking.

This interpretation is consistent with a growing body of literature showing that LLMs exhibit strong competence in generating coherent syntactic-semantic structures, which can be perceived as cognitively meaningful even in the absence of underlying experiential processes [4,22].

### Structural Differences Among Mentalization Dimensions

The axis-by-axis analysis reveals a differentiated profile of performance across MBT dimensions.

The implicit-explicit dimension and synthetic profile showed the highest scores, indicating that the model is particularly effective in explicating mental connections and producing coherent narrative syntheses. This pattern is consistent with previous findings suggesting that LLMs tend to favor linguistic transparency and structured organization of discourse [22].

The self-other axis confirms the model's ability to reproduce the linguistic structures associated with perspective differentiation, consistent with the grammar of intersubjective perspective-taking [10]. However, this differentiation remained predominantly descriptive rather than affectively grounded.

In contrast, the cognitive-affective dimension—and particularly the internal-external axis—showed the lowest scores, highlighting a difficulty in integrating internal states with contextual dynamics.

Generated profiles tended to describe subjective experience and external context in parallel, rather than as part of a dynamic or causal structure. This limitation is consistent with previous findings indicating that LLMs have reduced capacity to represent the situated relationship between mind and environment [8].

## Linguistic Coherence and Experiential Depth

Clinical evaluations converge in attributing high linguistic coherence to the generated profiles. The discourse is regular, logical connections are explicitly articulated, and perspectival elements are structured according to patterns that favor perceptions of cognitive reflectivity—consistent with the association between syntactic order and perceived mentalizing described by Frith and Frith [10].

However, this formal solidity is not accompanied by experiential depth comparable with human mentalization.

Clinicians consistently noted affective neutrality, describing emotional content as linear, predictable, and lacking the ambivalence, tonal oscillations, and regulatory variations typical of embodied mentalization.

This pattern is consistent with the nonsituated nature of language models, which lack the embodied experience required to generate fully integrated emotional representations [12] and tend to produce affectively plausible but contextually unanchored content [8]. The result is a form of ordered but nonexperiential reflectivity—the model integrates perspectives and mental connections at a formal level, without clear evidence of the emotional dynamics typically associated with human mentalization. This limitation should not be interpreted as a failure, but as a structural constraint rooted in the absence of phenomenological experience.

Accordingly, the observed coherence reflects clinician-perceived interpretability within a linguistic and evaluative framework, rather than evidence of clinical validity, diagnostic accuracy, or therapeutic effectiveness.

## A Form of Algorithmic Reflectivity

Within this empirical framework, the model's behavior can be described as a form of algorithmic reflectivity, understood as a modality of representing mental states based solely on the statistical organization of language.

From a methodological standpoint, the distinction between linguistic form and mental process is essential, as it underlies numerous interpretative errors [5].

The texts produced show syntactic and semantic coherence and integrate multiple perspectives according to ordered schemes recognizable by clinicians as compatible with cognitive reflectivity [10].

However, this formal coherence does not necessarily imply intentional or affective processes. Emotional representation remains descriptive and linear, lacking the ambivalence, discontinuity, and regulatory dynamics that characterize human mentalization [1].

This limitation may be consistent with accounts emphasizing the lack of embodiment in language models, which do not possess the bodily experience required to generate situated internal states or contextual affective variations [8,12].

Within the present framework, algorithmic reflectivity may be understood as a capacity for formal ordering and integration of discourse, distinct from psychological mentalization, which requires phenomenological and relational experience beyond the reach of linguistic systems.

## Algorithmic Reflectivity, Pseudoempathy, and Epistemic Trust

In this study, algorithmic reflectivity is defined as an output-level phenomenon. Clinicians' perception that generated profiles display structurally coherent and clinically recognizable mentalizing language. This construct remains deliberately agnostic regarding internal mental states and is best understood through the distinction between formal linguistic competence and experiential understanding [5].

This distinction becomes clinically relevant when contrasted with pseudoempathy, defined as the impression of being understood and emotionally recognized in interaction. While coherent and structured language may generate reassurance or perceived attunement, it does not in itself establish the relational and epistemic conditions required for genuine therapeutic change [13,23].

Epistemic trust, in this context, is not a property of the system, but an emergent relational phenomenon grounded in embodied interaction, affective contingency, and the capacity to tolerate uncertainty.

Within this framework, the difference between the 2 constructs becomes clear. Pseudoempathy concerns relational impression, whereas algorithmic reflectivity concerns the structural organization of discourse.

Unlike pseudoempathy, which primarily describes a phenomenological and interactional effect, algorithmic reflectivity is introduced here as a structurally grounded and operationalizable construct, assessed through observable linguistic organization and clinician-rated coherence. This allows empirical evaluation of mentalization-like properties at the level of discourse.

Conflating these constructs risks attributing experiential and relational depth to what may instead be a formally coherent but nonexperiential linguistic structure.

A particularly relevant finding concerns the inclusion of regulatory functions as a distinct dimension of analysis. While the other dimensions primarily capture structural and representational aspects of mentalization, the regulatory domain reflects the capacity to modulate affective states, tolerate ambiguity, and integrate potentially conflicting mental contents.

From a clinical perspective, this dimension occupies a pivotal position, as it is closely related to the capacity to maintain reflective functioning under conditions of emotional arousal. Although the model generated linguistically coherent outputs in this domain, high regulatory scores should be interpreted cautiously. Rather than indicating genuine affect regulation, they may reflect the reproduction of linguistically encoded patterns of regulation.

This further supports the distinction between algorithmic reflectivity and experiential mentalization.

## Safety Considerations: Sycophancy and Amplification Risk

The presence of structurally coherent and clinically interpretable language may also entail risks when applied to vulnerable populations. LLMs have been shown to exhibit forms of "sycophancy," namely a tendency to align with or reinforce user-provided perspectives, even when these may be distorted or maladaptive [24,25].

In clinical contexts, such alignment may unintentionally amplify rigid beliefs, maladaptive narratives, or delusion-consistent interpretations if not embedded within appropriate professional supervision.

Therefore, while high perceived coherence represents a necessary condition for research or training applications, it is not sufficient to guarantee clinical safety.

Future research should investigate the boundary conditions under which algorithmic reflectivity may support reflective functioning versus reinforcing maladaptive cognitive or affective patterns.

## Theoretical Implications

The findings clarify the distinction between mentalization as a function (experiential, embodied, and affectively modulated) and mentalization as a linguistic form (descriptive, explicit, and structured).

The model's performance indicates that some components of mentalization, particularly those relating to cognitive perspective-taking, synthesis, and explicit inferential connections, can be adequately simulated through linguistic regularities.

In contrast, other components—contextual integration, affect regulation, and ambivalence—remain dependent on psychological processes not reducible to linguistic form.

This contributes to ongoing debates on the limits and possibilities of LLMs as tools for evaluating or simulating sociocognitive processes.

## Epistemological Implications

The emergence of algorithmic reflectivity raises important epistemological questions about the status of mental knowledge generated by LLMs.

First, the results show that the model's outputs can be interpreted as reflecting nonexperiential forms of knowledge representation and derive from the extraction of linguistic regularities rather than from subjective experience.

Thus, the model does not "know" mental states in the psychological sense; rather, it reconstructs them based on statistically learned linguistic schemas.

This gives rise to an epistemic distinction between:

- generative knowledge (reconstruction of mental states through linguistic patterns)
- phenomenological knowledge (lived experience, foundational to human mentalization)

Second, the results suggest that what clinicians recognize as mentalization is tied to the linguistic form of reflectivity, not to its mental origin.

This raises a further epistemic issue—clinical evaluation is sensitive to linguistic structure regardless of the presence or absence of genuine intentional experience.

Finally, the findings contribute to discussions on the epistemic nature of nonsituated systems; their capacity to generate mentalizing statements does not imply access to intentional states, but introduces a form of knowledge that is formal, nonintentional, nonexperiential, yet relevant to modeling aspects of reflective thinking.

This invites a reconsideration of the relationships among language, intentionality, and psychological knowledge in artificial systems.

## Limitations of the Study

This study presents several limitations that define the interpretative scope of the findings.

The results reflect expert-rated textual coherence and clinical interpretability rather than clinical validity, diagnostic utility, or therapeutic effectiveness. The design does not allow inferences regarding the model's internal states or the presence of genuine mentalizing processes.

Qualitative observations reported in the Results and Discussion sections were derived from informal clinician comments and were not analyzed using a formal qualitative methodology (eg, thematic analysis or structured coding). Accordingly, interpretative claims regarding affective neutrality, experiential depth, or emotional dynamics should be considered exploratory and hypothesis-generating.

Algorithmic reflectivity is intrinsically dependent on the quality of human input. Participants' reflective functioning was not psychometrically assessed, and variability in the depth and organization of responses may have influenced the coherence of the generated profiles. This limits generalizability to clinical populations characterized by impaired, fragmented, or prementalizing discourse.

The data structure is hierarchical, with dialogues nested within participants and story prompts. Although dialogue-level aggregation partially mitigates this issue, statistical dependencies may affect the independence of observations and the precision of descriptive estimates. Accordingly, all reported estimates should be interpreted as descriptive dialogue-level indicators rather than population-level inferences. Future studies should apply mixed-effects modeling approaches to explicitly account for clustering at the participant and stimulus levels.

Raters were not informed that the profiles were AI-generated, and no formal manipulation check was conducted. Therefore, it cannot be excluded that stylistic features influenced implicit assumptions regarding authorship. Such assumptions may have affected coherence ratings by shaping expectations about the plausibility, linguistic organization, or interpretative quality of the texts. In the absence of a posttask assessment, it was not possible to quantify raters' ability to discriminate between AI-generated and human-written texts, nor to determine whether such assumptions systematically influenced coherence ratings. As a result, the independence of plausibility judgments from potential source attribution remains unverified. Future studies should incorporate explicit discrimination tasks (eg, source classification accuracy, confidence ratings, and area under the curve metrics) to assess the extent to which perceived coherence is affected by beliefs about authorship.

Score distributions in the main condition were concentrated in the upper range, suggesting a potential ceiling effect that may have attenuated ICC estimates where between-dialogue variance was restricted. However, the presence of a negative control condition yielding consistently low scores indicates that the rating scale retained sufficient sensitivity across its full range. Accordingly, parametric summaries for ordinal Likert-scale data should be interpreted with caution.

The study was not designed to include a direct comparison with clinician-generated MBT profiles, as its primary aim was to assess the internal coherence and interrater stability of model-generated outputs rather than to benchmark them against expert production. Consequently, the findings of this study should not be interpreted as evidence of equivalence between model-generated and clinician-authored profiles. Rather, they indicate that the perceived coherence of generated profiles depends critically on their alignment with the originating dialogue. Future research should extend this design by incorporating clinician-written baselines and applying paired ordinal statistical tests (eg, Wilcoxon signed-rank) with effect size estimation to further contextualize model performance.

Finally, the absence of a human-generated baseline limits interpretability. The findings should therefore be considered exploratory and hypothesis-generating, rather than evidence of equivalence between human- and model-generated mentalization.

## Future Directions

The findings of this study open several unexplored research avenues that may contribute to a more precise understanding of the relationships among language, mentalization, and computational modeling of sociocognitive processes.

The first direction concerns the analysis of temporal dynamics through which the model progressively constructs reflectivity over the course of dialogue.

The analysis of final profiles does not capture how the model manages extended reflective sequences, possible disruptions in narrative continuity, or spontaneous variations in coherence across turns.

Methods from computational linguistics—such as thematic progression analysis, clustering of recurring semantic patterns, or multiturn perspectival flexibility metrics—could reveal evolutionary patterns in generated reflectivity.

A second area concerns the development of computational metrics for linguistic mentalization.

Quantitative indicators related to syntactic coherence, inferential depth, explicitation of mental connections, and perspectival differentiation could be systematically correlated with the clinical evaluations provided by MBT experts.

The creation of annotated datasets and validated linguistic markers would enable the development of objective tools for automated assessment of reflective thinking.

A third direction involves designing more ecological experimental tasks, able to test not only explicit reflectivity but also the preliminary phases of the mentalizing process.

In particular, scenarios eliciting prementalizing states—teleological stance, psychic equivalence, and pretend mode—could assess how the model handles conditions corresponding to early and less mature forms of mental representation.

Dialogues with high affective intensity, concrete thinking, or pseudomentalizing dynamics would make it possible to observe whether the LLM maintains formal coherence, tends to reinterpret experience in mentalized terms, or reflects linguistic signals characteristic of prementalizing modes.

Such tasks would assess not only the model's reflective capacity but also its ability to respond appropriately to states in which mentalization is fragile, discontinuous, and marked by recurrent breakdowns.

Finally, language models could serve as computational baselines for investigating the linguistic structure of human reflective thought.

An LLM can provide a formal reference useful for isolating which components of mentalization emerge from subjective experience and which derive from discursive regularities.

Moreover, it can generate controlled narrative variants that allow researchers to study how clinicians perceive reflectivity across different contexts, offering an experimental tool for analyzing the sensitivity of clinical judgment.

These directions confirm that, despite lacking any form of psychological experience, language models can constitute a computational laboratory for investigating the linguistic grammar of mentalization, including early prementalizing modes.

A further priority for future research is the inclusion of a human-generated baseline, which would allow more direct comparison between model-produced and clinician-produced mentalization profiles and improve interpretability of the findings.

Future research may thus refine clinical tools, enrich theoretical understanding, and expand the possibilities for integration among cognitive science, psychotherapy, and AI.

## Conclusions

The results of this study indicate that, within the present experimental design, model-generated profiles are perceived by clinicians as meeting formal criteria associated with reflective discourse.

Mean scores and interrater agreement values indicate that in dimensions most closely linked to the cognitive structuring of language—such as the explication of mental connections, perspectival differentiation, and synthetic organization—the model produces stable and clinically interpretable outputs.

These results reflect clinician-rated coherence and interpretability within the study design, and do not constitute evidence of clinical validity, diagnostic utility, or therapeutic effectiveness.

These findings suggest that some components of mentalization, particularly those tied to linguistic form and discourse logic, may be approximated through statistical regularities in language.

At the same time, lower scores observed in axes related to the integration of internal states and external context, and to affective modulation, highlight the model's structural limitations in representing elements of experience that extend beyond linguistic structure.

These areas require the ability to situate emotional experiences within a dynamic, contextually embedded frame—an ability intrinsically tied to subjective experience and thus not fully modeled by systems lacking embodiment.

Taken together, these results delineate a clear distinction between what an LLM can simulate and what remains unique to human mentalization. The model appears capable of reproducing several formal linguistic characteristics associated with reflective thought, without providing evidence of the full range of psychological dynamics typically associated with human mentalization.

These findings suggest a potential role for language models in research and training contexts, particularly for the analysis of the linguistic components of mentalization.

At the same time, the findings do not support the interpretation of such systems as autonomous tools for clinical intervention. Rather, they indicate that the model's outputs lack the experiential, affective, and regulatory dimensions that constitute the core of the mentalizing function.

In conclusion, the study shows that LLMs offer a specific contribution: the capacity to render observable the formal and grammatical structure of reflective discourse, allowing the isolation and analysis of its linguistic components.

By contrast, the human component remains indispensable for affective experience, contextual grounding, and emotional regulation.

Understanding this distinction is crucial for defining the role of language models in clinical, educational, and research contexts and represents a starting point for future explorations aimed at clarifying the boundaries between linguistic simulation and real psychological processes.

### Acknowledgments

The authors would like to thank the clinicians who participated as raters in this study for their time, expertise, and careful evaluations. The authors also acknowledge the human participants who contributed written responses for research purposes. No financial compensation was provided to participants or raters.

The authors further acknowledge the use of generative artificial intelligence tools to support the translation of the manuscript into English. All translated content was carefully reviewed and validated by the authors to ensure accuracy and consistency with the original text.

### Funding

This study was funded by the Digital Transformation Institute, an independent non-governmental organization engaged in research activities on artificial intelligence and digital transformation. The funding body had no role in the study design, data collection, data analysis, interpretation of results, manuscript preparation or the decision to submit the manuscript for publication.

### Data Availability

The aggregated clinician rating dataset supporting the findings of this study is publicly available within the supplementary materials (Multimedia Appendix 3).

The underlying anonymized dialogue transcripts and qualitative textual materials are not publicly available due to ethical and confidentiality considerations related to participant-generated content. These materials may be made available from the corresponding author upon reasonable request for research purposes and subject to appropriate data protection safeguards.

### Authors' Contributions

Conceptualization: SE, GC
Methodology: SE, GC, LK, ES-S

Formal analysis: GR
Investigation: LK, ES-S
Supervision: SE, GC
Writing – original draft: SE
Writing – review & editing: SE, GC, GR, LK, ES-S

## Conflicts of Interest

SE, GC, and GR are affiliated with the Digital Transformation Institute, which provided funding for this study. The authors declare that this affiliation did not influence the study design, data analysis, interpretation of results, or the conclusions drawn. All other authors have no competing interests to declare.

## Multimedia Appendix 1

Full system prompt, prompt architecture, and inference parameters.
[DOCX File (Microsoft Word File), 43 KB-Multimedia Appendix 1]

## Multimedia Appendix 2

ICC(2,k)-ICC(3,1) details.
[XLSX File (Microsoft Excel File), 12 KB-Multimedia Appendix 2]

## Multimedia Appendix 3

Raters evaluation dataset.
[XLSX File (Microsoft Excel File), 27 KB-Multimedia Appendix 3]

## Abbreviations

**ICC:** intraclass correlation coefficient
**LLM:** large language model
**MBT:** mentalization-based treatment
**NLP:** natural language processing